\PassOptionsToPackage{table}{xcolor}
\documentclass[runningheads]{llncs}

\usepackage{eccv}

\usepackage{eccvabbrv}

\usepackage{graphicx}
\usepackage{booktabs}
\usepackage{graphicx}
\usepackage{amsmath}
\usepackage{amssymb}
\usepackage{physics}

\usepackage{multirow}
\usepackage{multicol}
\usepackage{amsmath}
\usepackage{color}

\usepackage{listings}
\usepackage{algorithm}
\usepackage{algpseudocodex}

\usepackage[accsupp]{axessibility}  

\usepackage{hyperref}

\usepackage{orcidlink}
\usepackage{pifont}

\newcommand{\calL}{\mathcal{L}}

\usepackage[normalem]{ulem}

\begin{document}
\title{MetaAug: Meta-Data Augmentation for Post-Training Quantization}

\titlerunning{MetaAug: Meta-Data Augmentation for Post-Training Quantization}

\author{Cuong Pham\inst{1} \and
Hoang Anh Dung \inst{1}
\and
Cuong C. Nguyen\inst{2} \and
Trung Le\inst{1} 
\and \\
Dinh Phung\inst{1,3}
\and
Gustavo Carneiro \inst{2}
\and
Thanh-Toan Do \inst{1}}

\authorrunning{Pham et al.}

\institute{Department of Data Science and AI, Monash University, Australia 
\and
Centre for Vision, Speech and Signal Processing, University of Surrey, UK 
\and
VinAI, Vietnam
\\
\email{\{cuong.pham1, hoang.dung, trunglm, dinh.phung, toan.do\}@monash.edu} \\
\email{\{c.nguyen, g.carneiro\}@surrey.ac.uk}
}
\maketitle

\begin{abstract}
  Post-Training Quantization (PTQ) has received significant attention because it requires only a small set of calibration data to quantize a full-precision model, which is more practical in real-world applications in which full access to a large training set is not available. However, it often leads to 
  overfitting on the small calibration dataset. Several methods have been proposed to address this issue, yet they still rely on only the calibration set for the quantization 
  and they do not validate the quantized model due to the lack of a validation set. In this work, we propose a novel meta-learning based approach to enhance the performance of post-training quantization.  
  Specifically, to mitigate the overfitting problem, instead of only training the quantized model using the original calibration set without any validation during the learning process as in previous PTQ works, in our approach, we both train and validate the quantized model using two different sets of images.  
  In particular, we propose a meta-learning based approach to jointly optimize a {transformation network} and a quantized model through bi-level optimization. The {transformation network} modifies the original calibration data and the modified data will be used as the training set to learn the quantized model with the objective that the quantized model achieves a good performance on the original calibration data.
  Extensive experiments on the widely used ImageNet dataset with different neural network architectures demonstrate that our approach 
outperforms the state-of-the-art PTQ methods. Code is available at \href{https://github.com/cuong-pv/MetaAug-PTQ}{this https URL}.
    \end{abstract}

\section{Introduction}
\label{sec:intro}

Deep neural networks (DNNs) have received a substantial amount of attention due to their state-of-the-art performance in various tasks. However, deploying these networks on resource-constrained devices is challenging due to the limited computational resources and memory footprint. To make DNNs more efficient, network quantization~\cite{han2015deep,BinaryConnect,nagel2019data,AdaRound,cai2020zeroq,xu2020generative} has been extensively studied due to its computational and storage benefits. Quantization is the process of reducing the precision of the weights and activations of DNNs.   
Depending on the available training data, network quantization can be divided into two main categories: quantization-aware training (QAT)~\cite{BinaryConnect,XNOR-Net,DSQ,yang2019quantization,defossez2021differentiable,LSQ,shin2023nipq} and post-training quantization (PTQ)~\cite{AdaRound,BRECQ,Wei2022QDropRD,liu2023pd,Genie}. Although QAT generally results in better performance compared to PTQ and can reduce the gap to full-precision accuracy for low-bit quantization, it requires a large training set to retrain DNNs on the targeting dataset.  
This may not be practical for many real-world applications where a large training dataset is unavailable or access to it is restricted due to security and privacy concerns. 

To tackle this problem, PTQ has been investigated
because it only employs a small calibration dataset to quantize a well-trained full-precision model. 
However, this approach often results in overfitting to the used small calibration set~\cite{zheng2022leveraging,Wei2022QDropRD,liu2023pd}. 
Various methods have been proposed to mitigate this overfitting issue. In QDrop \cite{Wei2022QDropRD}, the authors propose to mitigate overfitting in PTQ by randomly dropping quantized activations. 
In \cite{zheng2022leveraging}, the authors utilize activation regularization by minimizing the difference between the intermediate features of the full-precision model and the quantized model. In PD-Quant \cite{liu2023pd}, the authors indicate the performance degradation in PTQ due to a severe overfitting on the calibration set and they also adopt activation regularization to counteract overfitting. In addition, they introduce activation distribution correction as regularization to further alleviate overfitting by encouraging the distributions of quantized activations to match the batch normalization (BN) statistics from the BN layers of the full-precision model. 
Although different strategies have been proposed, they all rely on the original calibration data for training the quantized model and they do not have a validation set to validate the quantized model during the quantization process. This could lead the quantized model to be prone to overfitting on the calibration set. 

Different from previous works~\cite{zheng2022leveraging,Wei2022QDropRD,liu2023pd} in PTQ that use the calibration set for training and do not have a validation set to validate the quantized model, in this work, we propose to perform the quantization using two different sets -- a modified version of the calibration set is used as the training data for learning the quantized model, while the original calibration data is used as the validation set to validate the quantized model. The modified data is produced by a learnable transformation network that takes the original calibration data as input. 
Our work aims to jointly optimize both the transformation network and the quantized network with the objective that they lead to a good performance of the quantized network on the validation set, i.e., the original calibration set. However, this aim is  nontrivial. This is because the problem is a nested optimization in which the optimization for the transformation network 
is to minimize a validation loss of the quantized network while the quantized network itself is subjected to another optimization 
with some quantization loss. 

To tackle this challenge, 
{we propose a novel meta-learning based PTQ approach} in which the transformation network and the quantized network are jointly optimized through a bi-level optimization. 
A noticeable challenge in this approach is the possibility of the transformation network to be degenerated into an identity mapping. Consequently, such scenario can result in overfitting in the quantization process, as the training and the validation of the quantized model use the same original calibration data.
To prevent this situation, we deeply investigate approaches to make the transformation network capable of preserving the 
information of the original calibration data while still giving it the flexibility to avoid 
being a trivial (i.e., an identity) transformation. Specifically, 
we investigate three different losses for semantic preservation, including a probabilistic knowledge transfer loss. This encourages the transformation network to capture the feature distributions of the original calibration data which consequently preserves the 
information of the calibration data. 
In addition, we also propose using a margin loss to discourage the transformation network from being a trivial transformation. 
We validate our proposed approach on the widely used ImageNet dataset with different neural network architectures by comparing it with state-of-the-art methods. The extensive empirical results demonstrate that 
our method outperforms the state-of-the-art PTQ methods. 

Our contributions can be summarized as follows:
    \ding{182 } We propose a novel meta-learning based method to mitigate the overfitting problem in PTQ. The proposed approach jointly optimizes a transformation network 
    and a quantized model. During the learning process, 
    the outputs of the transformation network and the original data are used for training and validating the quantized model, respectively.  
    To the best of our knowledge, this is the first work that tackles the overfitting problem in PTQ through a meta-learning bi-level optimization approach.
    \ding{183 } 
    We investigate different losses for training the transformation network such that the outputs of the network preserve the feature 
    information of the original calibration data.
    Furthermore, we also propose using a margin loss to discourage the transformation network from being an identity mapping. 
    \ding{184 } We validate our proposed approach on the widely used ImageNet dataset across various neural network architectures. Extensive experiments demonstrate that the proposed method outperforms the state-of-the-art PTQ methods. 

\section{Related work}
\label{sec:related_work}
\paragraph{\textbf{Uniform quantization.}}
To uniformly quantize a tensor $w$ to $b$ bit-width, the support space is uniformly discretized into $2^b -1$ even intervals. As a result, the original 32-bit single-precision value is mapped to an unsigned integer within the range of [0, $2^b-1$], or a signed integer within the range of [$-2^{b-1}$, $2^{b-1}-1$]. 
Supposed that $Q_b$ is the quantization function with a bit-width of \(b\), the quantization function $Q_b$ is defined as follows:
\begin{equation}
    \hat{w} = Q_b(w;s) = s \times \operatorname{clip} \left( \left\lfloor \frac{w}{s} \right\rceil, n, p \right),
\end{equation}
where \(s\) represents the scaling factor, $\lfloor . \rceil$ denotes the rounding-to-nearest function, and clip() represents the clipping function.
For unsigned data (e.g., activations with ReLU or Sigmoid)
$n=0$, $p=2^b-1$, and for signed data (e.g., weights) $n=-2^{b-1}$, $p=2^{b-1}-1$. 
In PTQ, rounding-to-nearest is the most common rounding function by minimizing the quantization error. However,  the most recent state-of-the-art approaches \cite{AdaRound,BRECQ,liu2023pd,Wei2022QDropRD,Genie} have shown that a learnable rounding function can improve the performance of quantized models. 
The quantization function $Q_b$ in those studies is defined as:
\begin{equation}
    \hat{w} = Q_b(w;s, v) = s \times \operatorname{clip} \left( \left\lfloor \frac{w}{s} \right\rfloor + h(v), n, p \right) \quad \text{s.t.: } v \in \{ 0, 1 \},
\end{equation}
where $h(v)$ is a learnable function that maps the value of $v$ to either 0 or 1. Note that during training, the scaling factor \(s\) is fixed in AdaRound~\cite{AdaRound}, while being learned simultaneously with the rounding function \(h(v)\) in Genie~\cite{Genie}. In our work, we adopt the Genie~\cite{Genie} approach for weight quantization and LSQ~\cite{LSQ} for activation quantization.

\paragraph{\textbf{Post training quantization (PTQ)}.}
This quantization approach has gained considerable attention recently because it does not require access to large amounts of data and can operate effectively with minimal or even unlabeled training data. 
This method is particularly useful when full access to training data is not possible. 
In addition, it is useful for large models that are not suitable for QAT due to their substantial training time. 
In AdaRound~\cite{AdaRound}, the authors propose using a learnable rounding function instead of the traditional rounding-to-nearest approach to quantize the model layer by layer. Based on this, BRECQ \cite{BRECQ} further improves the performance of PTQ by proposing block reconstruction (e.g., 4 blocks in ResNet18 \cite{resnet}) that considers the dependency of layers' outputs in each block of the neural network. In~\cite{ma2023solving}, the authors address the problem of oscillation in PTQ. They propose a method to identify blocks within a network that should be jointly optimized and quantized. 
In QDrop~\cite{Wei2022QDropRD}, their framework exploits a mechanism randomly dropping quantized activations to improve the flatness of the quantized model. 
In~\cite{zheng2022leveraging}, an activation regularization is proposed, by minimizing the difference between the intermediate features after the activation function of the quantized model and the full-precision model. 
In addition to activation regularization, another method named PD-Quant \cite{liu2023pd} demonstrates performance improvement by correcting the feature distribution of calibration data to follow the feature distribution of full-training data based on batch normalization (BN) statistics from the BN layers of the full-precision model. In Genie~\cite{Genie}, the authors propose to learn the scale and rounding functions simultaneously to further improve the performance of PTQ. Another approach for PTQ is Bit-Shrinking~\cite{lin2023bit}, which incorporates sharpness-aware minimization into the quantization process. In that method, the authors suggest progressively reducing the bit-width of quantized models to limit the instantaneous sharpness of the objective function. 
It is worth noting that all the mentioned methods only rely on the original calibration data for training the quantized model. They do not have a validation set to validate the quantized model during the quantization process. This could lead the quantized model to be prone to overfitting on the calibration set.

\paragraph{\textbf{Meta-learning.}}
Meta-learning methods can be divided into three categories: optimization-based, model-based, and metric-based methods. 
Optimization-based meta-learning investigates the optimization in the task adaptation step and uses training tasks to improve that optimization (e.g., learn a good learning rate~\cite{li2017meta}, model initialization~\cite{finn2017model}, updating rule~\cite{ravi2017optimization} or even a data-driven optimizer~\cite{andrychowicz2016learning}).
Among many optimization-based meta-learning methods, MAML~\cite{finn2017model} is one of the most popular ones. MAML aims to learn a meta-model that can quickly adapt to new tasks with few training examples. 
Since then, many variants of this optimization-based approach have been proposed to further enhance the performance \cite{sharp-maml,sign-maml,fo-maml,mix-maml}. 
 On the other hand, model-based meta-learning models, such as Memory-Augmented Neural Networks (MANNs)\cite{mann} and Recurrent Meta-Learners framework\cite{Duan2016RL2FR,Wang2016LearningTR}, maintain an internal representation of a task during training. This internal state is periodically updated based on new inputs and makes great contribution to the model output.
Finally, the last branch of meta-learning methods -- metric-based frameworks\cite{Vinyals2016MatchingNF,Snell2017PrototypicalNF,Satorras2017FewShotLW,Sung2017LearningTC,Koch2015SiameseNN}, are designed to learn an embedding function to map all data points to a metric embedding space. 
Overall, despite demonstrating the ability to generalize the model over unseen data, there is still not enough attention regarding the applicability in PTQ of meta-learning.
\paragraph{\textbf{Meta-Learning for Network Quantization.}} 
Several works~\cite{chen2019metaquant,Wang2020AutomaticLH,Youn2022BitwidthAdaptiveQN,Kim2023MetaMixMP} have utilized meta-learning for quantization. In MetaQuantNet \cite{Wang2020AutomaticLH}, the authors propose a framework that can automatically search for the best quantization policy with meta-learning before using that policy for quantization. On the other hand, MEBQAT\cite{Youn2022BitwidthAdaptiveQN} attempts to leverage the meta-learning mechanism to optimize a mixed-precision quantization model capable of adapting to different bit-width scenarios quickly without hurting the model's performance. Another work named MetaMix \cite{Kim2023MetaMixMP} points out the activation instability problem in existing methods for mixed-precision quantization and aims to tackle this problem with meta-learning. 
However, these methods focus on quantization-aware training, while our work focuses on mitigating overfitting in post-training quantization. Additionally, 
all of these methods only leverage meta-learning to improve their quantization mechanisms without considering the impact of calibration data to their frameworks. 
To the best of our knowledge, our work is the first to leverage meta-learning in the context of post-training quantization, from the perspective of data optimization.

\section{Proposed method}
\label{sec:proposed}

\subsection{Meta-learning formulation for PTQ}
Let $S =\{x_i\}_{i=1}^N$ be the calibration set.
Given a sample $x_i \sim S$, consider a full-precision model $\theta_{\mathrm{FP}}$ and a quantized model $\theta_Q$, our objective is to learn a transformation network $T$ that modifies the calibration sample $x_i$ into an adaptive sample $T(x_i)$ beneficial for model generalization.  
The data sample $T(x_i)$ outputted by $T$ is then utilized to optimize the quantized network $\theta_Q$ to get a model $\widehat{\theta}_Q$ after a number of gradient descent steps. The optimal $T$ is then determined based on performance of the model $\widehat{\theta}_Q$ on the original data set $S^{v} =\{x_i^{v}\}_{i=1}^N$ (in this context, $S^{v} = S$). The bi-level objective function is defined as:
\begin{align}
    T^{*} & = \arg\min_{T} \frac{1}{N} \sum_{i=1}^{N} \mathcal{L}_{\mathrm{val}} ( \widehat{\theta}_{\mathrm{Q}}, x_i^{v}) \label{eq:opt_T_alpha} \\
    & \hspace{-0.4cm} \text {s.t.: } \widehat{\theta}_{\mathrm{Q}}= \arg \min_{\theta_Q} \frac{1}{N} \sum_{i=1}^{N} \mathcal{L}_Q(\theta_{\mathrm{Q}}, T(x_i)). \label{eq:opt_theta_Q}
\end{align}
The objective function in \cref{eq:opt_T_alpha} represents a bi-level optimization problem, typically solved in two stages. The first stage presented in \cref{eq:opt_theta_Q} involves optimizing $\theta_Q$ using the modified data $\{T(x_i)| i=1,2,..,N\}$. This stage can be addressed using gradient-based optimization methods, such as SGD or Adam, as follows:
\begin{equation}
    \widehat{\theta}_{\mathrm{Q}} = \theta_{\mathrm{Q}}  - \frac{\eta}{N} \sum_{i=1}^{N} \grad_{\theta_{\mathrm{Q}}} \mathcal{L}_Q(\theta_{\mathrm{Q}}, T(x_i)),
\end{equation}
where \(\eta\) is the learning rate to update \(\widehat{\theta}_{\mathrm{Q}}\).

The second stage involves updating $T$ based on the model $\widehat{\theta}_{\mathrm{Q}}$, focusing on the performance evaluated on the original data $x_i^v$. This update corresponds to the upper-level optimization. This can be expressed as follows:
\begin{equation}
    \label{eq:opt_T_alpha2}
    T \gets T - \frac{\gamma}{N} \sum_{i=1}^{N} {\grad_{T} \mathcal{L}_{\mathrm{val}}(\widehat{\theta}_{\mathrm{Q}}, x_i^v)}
\end{equation}
where $\gamma$ is the learning rate to update $T$.

As shown in~\cref{eq:opt_T_alpha2}, optimizing $T$ requires calculating the gradient of the validation loss $\mathcal{L}_{\mathrm{val}}(\widehat{\theta}_{\mathrm{Q}}, x_i^v)$ with respect to $T$. 
Using the chain rule, the computation can be performed as follows:

\begin{equation}
    \label{eq:grad_T_alpha1}
    \begin{aligned}[b]
        \grad_{T} \mathcal{L}_{\mathrm{val}} \left( \widehat{\theta}_{\mathrm{Q}}, x_{i}^{v} \right) & = \grad_{T}^{\top} \widehat{\theta}_{\mathrm{Q}} \times \grad_{\widehat{\theta}_{\mathrm{Q}}} \mathcal{L}_{\mathrm{val}}(\widehat{\theta}_{\mathrm{Q}}, x_{i}^{v}) \\
        & = \grad_{T}^{\top} \left[ \theta_{\mathrm{Q}}  - \frac{\eta}{N} \sum_{j=1}^{N} \grad_{\theta_{\mathrm{Q}}} \mathcal{L}_Q(\theta_{\mathrm{Q}}, T(x_j)) \right] \times \grad_{\widehat{\theta}_{\mathrm{Q}}} \mathcal{L}_{\mathrm{val}}(\widehat{\theta}_{\mathrm{Q}}, x_{i}^{v}) \\
        & = - \frac{\eta}{N} \grad_{T}(\sum_{j=1}^{N}\grad_{\theta_{\mathrm{Q}}} \mathcal{L}_{\mathrm{Q}} (\theta_{\mathrm{Q}}, T(x_{j})))^{\top}
        \times \grad_{\widehat{\theta}_{\mathrm{Q}}} \mathcal{L}_{\mathrm{val}}(\widehat{\theta}_{\mathrm{Q}}, x_{i}^{v}),
    \end{aligned}
\end{equation}
where \({}^{\top}\) denotes the transpose operator.

\paragraph{\textbf{Regarding} $\mathcal{L}_Q$ \textbf{in} (\ref{eq:opt_theta_Q}).}
We adopt the block-wise~\cite{BRECQ} quantization method to sequentially quantize the full-precision model $\theta_{\mathrm{FP}}$ to get the quantized model $\theta_{\mathrm{Q}}$. Given the pre-trained full-precision model $\theta_{\mathrm{FP}}$ consisting of $L$ blocks, we sequentially quantize the model block by block and update the transformation network $T$ to minimize the validation loss of $\widehat{\theta}_{\mathrm{Q}}$ on the original calibration data $S$.
The loss in~\cref{eq:opt_theta_Q} updating the ${l}^{th}$ block of the model $\theta_{\mathrm{Q}}$ to obtain the model $\widehat{\theta}_{\mathrm{Q}}$ is defined as follows:
\begin{equation}
    \label{eq:loss_Q_1}
    \begin{aligned}
        \mathcal{L}_Q(\theta_{\mathrm{Q}}, T(S)) = \frac{1}{N} \sum_{i=1}^{N} \norm{ A_{FP}^{l}(T(x_i)) - A_Q^{l}(T(x_i)) }^2, \\
    \end{aligned}
\end{equation}
 where $A^{l}(x_i)$ and and $A_Q^{l}(x_i)$ are the activations of the ${l}^{th}$ block of the full-precision model $\theta_{\mathrm{FP}}$ and the quantized model $\theta_{\mathrm{Q}}$ for sample $T(x_i)$, respectively.

 \paragraph{\textbf{Regarding} $\mathcal{L}_{\mathrm{val}}$ \textbf{in} (\ref{eq:opt_T_alpha}).}
As shown in~(\ref{eq:opt_T_alpha}), our goal is to 
minimize the validation loss of $\widehat{\theta}_{\mathrm{Q}}$ on the original data. Therefore, at the validation step, we validate the quantized model $\widehat{\theta}_{\mathrm{Q}}$ on the original calibration set $S = \{x_i\}_{i=1}^{N}$. We adopt Kullback-Leibler divergence loss to validate the quantized model $\widehat{\theta}_{\mathrm{Q}}$ on the original calibration data $S$, which is defined as follows:
\begin{equation}
    \label{eq:loss_Q}
    \begin{aligned}
        \mathcal{L}_{\mathrm{val}}( \widehat{\theta}_{\mathrm{Q}}, S) = \frac{1}{N} \sum_{i=1}^{N} \operatorname{KL} \left[\sigma( f_{\theta_{\mathrm{FP}}}(x_i)) \Vert \sigma(f_{\widehat{\theta}_{\mathrm{Q}}}(x_i)) \right], \\
    \end{aligned}
\end{equation}
where \(f\) is output of the model of interest and $\sigma(.)$ denotes the softmax operator.
\subsection{Transformation $T$ and regularizations to the modified images}
In this section, we discuss the definition of transformation network $T$ and objective functions to update $T$. 
The transformation network could be parameterized by an autoencoder, a UNet, or any other transformation network. In this work, we use the UNet~\cite{Unet} as the transformation network. The UNet is a widely used architecture for image-to-image translation tasks, consisting of an encoder and a decoder. The encoder is used to extract features from the input image, and the decoder is used to generate the output image. The UNet model has advantages over other autoencoders in retaining the fine feature information of the input image because it includes residual connections between the encoder and decoder. 
On the one hand, we expect the generated images $T(x_i)$ to retain the information of original images $x_i$. On another hand, the  transformation network $T$ should not be degenerated into a trivial solution i.e., an identity mapping, as it would have no effect on overfitting reduction. We investigate different objective functions to update $T$.
\subsubsection{Information Preservation.}
Given the original calibration set $S$, we have a corresponding transformed image set $S^{(g)} = \{T(x_i)|x_i
 \sim S, 1 \leq i \leq N\}$.
To transfer the information of images from $S$ to the generated set $S^{(g)}$, we investigate different losses for this purpose including a Mean Square Error loss (MSE), a Kullback–Leibler (KL) divergence loss, and a distribution preservation loss. 
The MSE between outputs from the full-precision model of original images and generated images is defined:
\begin{equation}
    \label{eq:MSE}
    \mathcal{L}_{\mathrm{MSE}}(T,S) =  \frac{1}{N} \sum_{i=1}^{N} \norm{ f_{\theta_{\mathrm{FP}}}(x_i) - f_{\theta_{\mathrm{FP}}}(T(x_i)) }^2.
\end{equation}

The KL loss between outputs from the full precision model of  original images and generated images is defined: 
\begin{equation}
    \label{eq:KL}
    \mathcal{L}_{\mathrm{KL}}(T,S) =   \frac{1}{N} \sum_{i=1}^{N} \operatorname{KL} \left[ \sigma(f_{\theta_{\mathrm{FP}}}(x_i)) \Vert \sigma(f_{\theta_{\mathrm{FP}}}(T(x_i)) \right]
\end{equation}

It is worth noting that the losses (\ref{eq:MSE}) and (\ref{eq:KL}) only consider pairwise distances between corresponding features from the full-precision (FP) and quantized models, without considering the 
information between samples. 
Therefore, we also consider another information preservation loss that aims to retain the whole dataset's distribution information by leveraging the distribution probabilistic loss that has been used in~\cite{PKT}. Specifically,
we first estimate the conditional probability density of any two data points within the feature space \cite{tsne,PKT}, which is formulated as:

\begin{equation}
    \label{eq:distribution_proba}
    \mathcal{P}_{i|j} = \frac{K(f_{\theta_{\mathrm{FP}}}(x_i),f_{\theta_{\mathrm{FP}}}(x_j))}{\sum_{\substack{k=1 \\ k \neq j}}^{N} K(f_{\theta_{\mathrm{FP}}}(x_k),f_{\theta_{\mathrm{FP}}}(x_j))},
\end{equation}

where $K(a, b)$ is a kernel function and $\mathcal{P}_{i|j}$ is the probability of $x_i$ given $x_j$.  Following PKT~\cite{PKT}, we adopt the cosine similarity metric $K(a, b)=\frac{1}{2}(\frac{a^Tb}{\norm{a}\norm{b}}+1)$ as kernel function.
To encourage the feature distribution matching between the original dataset $S$ and the generated dataset $S^{(g)}$, original image $x_i$ should share the same probability distribution with its corresponding generated image $T(x_i)$, so the distribution preservation loss $\mathcal{L}_{DP}$ is defined as:
\begin{equation}
    \label{eq:PKT}
    \mathcal{L}_{DP}(T,S) =  \frac{1}{N}\sum_{i=1}^{N} \operatorname{KL} \left[ \mathcal{P}_i \Vert \mathcal{P}^{(g)}_i \right],
\end{equation}
where $\mathcal{P}_i$ and $\mathcal{P}^{(g)}_i$ are the conditional probability distributions of the extracted features from the full precision model of original image $x_i$ and generated image $T(x_i)$, respectively. 

\paragraph{\textbf{Identity Prevention.}} To encourage the transformation network not to be an identity, we propose using the following loss:
\begin{equation}
    \label{eq:margin}
\mathcal{L}_{\mathrm{margin}}(T,S) = \frac{1}{N} \sum_{i=1}^{N} \max \left( 0, \epsilon - \frac{1}{M}\norm{x_i - T(x_i)}^2 \right),
\end{equation}
where \(\epsilon\) is a threshold to encourage 
that the difference between the generated data and the original data is not lower than the threshold, and M is the total number of pixels of image $x_i$. 

\paragraph{\textbf{Overall loss for training} $T$\textbf{.}} Combine objective loss in~\cref{eq:loss_Q}, and in~\cref{eq:margin} with either objective losses in~\cref{eq:MSE}, ~\cref{eq:KL},~\cref{eq:PKT}, we have the final combination loss to update $T$ as follows:
\begin{equation}
    \label{eq:loss_T_alpha_finalLP}
    \mathcal{L}_{T}(T,S) = \lambda_1 \mathcal{L}_{\mathrm{val}}( \widehat{\theta}_{\mathrm{Q}}, S) + \lambda_2\mathcal{L}_{\mathrm{margin}}(T,S) + \lambda_3 \mathcal{L}_{*}(T,S), 
\end{equation}
where $\mathcal{L}_{*} \in \{\mathcal{L}_{\mathrm{MSE}}, \mathcal{L}_{\mathrm{KL}}, \mathcal{L}_{\mathrm{DP}}\}$ and \(\lambda_{1}\), \(\lambda_{2}\) and \(\lambda_{3}\) are hyper-parameters.

The overall algorithm of our proposed method is presented in \cref{alg:meta_ptq_real}.
\begin{algorithm}[t!]
    \caption{Data modification for post-training quantization.}
    \label{alg:meta_ptq_real}
    \begin{algorithmic}[1]
        \Procedure{Train}{$\theta_{\mathrm{FP}}$,$S$}

            \LComment{$\theta_{\mathrm{FP}}$: weight of the full-precision model.}

            \LComment{$L$: Number of blocks in the full-precision model.}

            \LComment{$S$: Calibration data.}

            \LComment{$N_{T}$: Number of iterations to update $T$.}
            \LComment{$N_Q$: Number of iterations to quantize model.}

            \LComment{$T$: Transformation network to modify calibration dataset $S$.}

            \State Initialize the quantized model $\theta_Q$ from $\theta_{FP}$ using LAPQ~\cite{nahshan2021loss}.
            \State Warm up the transformation network $T$.
            \For{$l = 1$ to $L$}
            \For{$t = 1$ to $N_{T}$}
            \State Sample a mini-batch: $\mathbb{B}=\left\{ x_i : x_i \sim \cal{S} \right\}$
                \State Modify $\mathbb{B}$ with the transformation network $T$ to get $T(\mathbb{B}) = 
                \{T(x_i)\}_{i=1}^{|\mathbb{B}|}$
                \LComment{Forward pass and update the quantized model using modified data.}
                \State Compute: $\mathcal{L}_Q(\theta_{\mathrm{Q}}, T(\mathbb{B})) = \frac{1}{|\mathbb{B}|}\sum_{i=1}^{|\mathbb{B}|}||A_{FP}^{l}(T(x_i)) - A_Q^{l}(T(x_i))||^2$ \Comment{\cref{eq:loss_Q_1}}\\
                
                    \State Update $\widehat{\theta}_Q$: $\widehat{\theta}_Q$ $\gets$  Adam($\mathcal{L}_Q(\theta_{\mathrm{Q}}, T(\mathbb{B})))$
                    \Statex
                    \LComment{Validate $\widehat{\theta}_Q$ on the original calibration data.}
                    \State Sample a mini-batch data: $\mathbb{B}^{v}=\left\{ x^{v}_i : x^{v}_i \sim \cal{S} \right\}$

                    
                    \State Compute: $\mathcal{L}_{T}(T,\mathbb{B}^{v})$ \Comment{\cref{eq:loss_T_alpha_finalLP}}
                    \Statex
                    \State Update 
                    $T$: $T \gets$ Adam($\mathcal{L}_{T}(T,\mathbb{B}^v)$)
                    
            \EndFor
            \Statex
            \LComment{Quantize $l^{th}$ block of $\theta_Q$ using the original calibration data $\mathcal{S}$ and modified data with 
            the learned $T$}.
            \For{$t = 1$ to $N_Q$}
                \State Sample a mini-batch: $\mathbb{B}_q=\left\{ x_{qi} : x_{qi} \sim \mathcal{S}_q = T(\cal{S}) \cup \cal{S}  \right\}$
                \State Compute: $\mathcal{L}_Q(\theta_{\mathrm{Q}}, \mathbb{B}_q) = \frac{1}{|\mathbb{B}_q|}\sum_{i=1}^{|\mathbb{B}_q|}||A_{FP}^{l}(x_{qi})) - A_Q^{l}(x_{qi})||^2$ \\
                \State Update: $\theta_Q \gets$ Adam$(\mathcal{L}_Q(\theta_{\mathrm{Q}}, \mathbb{B}_q))$
            \EndFor
            \Statex
            \EndFor
            \State \Return quantized model $\theta_{\mathrm{Q}}$ and $T$.
        \EndProcedure
    \end{algorithmic}
\end{algorithm}

\section{Experiments}
\label{sec:experiments}
\subsection{Experimental setup}

\paragraph{\textbf{Datasets and network architectures.}}
We validate the proposed method on the ImageNet dataset~\cite{imagenet}. Following previous PTQ works~\cite{AdaRound,BRECQ,liu2023pd,Wei2022QDropRD,Genie,lin2023bit}, the calibration set used for training quantized models contains 1,024 images from the training set of the ImageNet dataset. The validation set of the ImageNet dataset containing 50,000 images is used as the test set.  
Following previous PTQ works~\cite{AdaRound,BRECQ,liu2023pd,Wei2022QDropRD,Genie,lin2023bit}, we evaluate our approach on the widely used network architectures including ResNet-18~\cite{resnet}, ResNet-50~\cite{resnet}, and MobileNetV2~\cite{Mobilenetv2}. 

\paragraph{\textbf{Implementation details.}} 
We utilize the UNet~\cite{Unet} as a transformation network to modify the calibration dataset. We use the Adam optimizer~\cite{Kingma2014AdamAM} with a learning rate of $5\times 10^{-6}$ to update the transformation network. This network is trained for 500 iterations with a batch size of 32. For the quantization of weights and activations, we follow current state-of-the-art approaches PTQ~\cite{AdaRound,BRECQ,liu2023pd,Wei2022QDropRD,Genie}. Specifically, for weight quantization, we learn both the scaling factor and rounding function following the Genie approach~\cite{Genie}. For activation quantization, we adopt the LSQ~\cite{LSQ}. We also keep the first and last layers at 8 bits as it does not increase much memory storage and helps prevent significant performance degradation~\cite{XNOR-Net}. The quantized model $\theta_Q$ is initialized from the full-precision model using LAPQ~\cite{nahshan2021loss} as previous works~\cite{AdaRound,BRECQ,liu2023pd,Wei2022QDropRD,Genie}. We use 
$2 \times 10^4$ iterations to quantize each block of the quantized model.  
When updating the transformation network $T$, to compute $\grad_{T} \mathcal{L}_{\mathrm{val}}$ in \cref{eq:grad_T_alpha1}, we utilize the \textit{higher}\footnote{https://github.com/facebookresearch/higher} library. 
We set the margin parameter $\epsilon$ in \cref{eq:margin} to 0.3 for experiments with ResNet-50, and 0.1 for experiments with ResNet-18 and MobileNetV2. We set the hyper-parameters $\lambda_{1}= 5$, and $\lambda_{2}=0.5$. We set $\lambda_{3}$ to 1, 5, and $3 \times 10^4$ for the $\calL_{MSE}$ in~\cref{eq:MSE}, $\calL_{KL}$ in~\cref{eq:KL}, and $\calL_{DP}$ in~\cref{eq:PKT}, respectively.
\begin{table}[t!]
            \centering
            \caption{Top-1 classification accuracy (\%) with the ResNet-18 architecture with different combinations of proposed losses evaluated on ImageNet dataset.}
            \label{tab:loss_combined}
  \centering
    \begin{tabular}{c|ccccc|c}
    \hline 
    \multirow{2}{*}{Setting} &\multirow{2}{*}{$\calL_{\mathrm{val}}$} & \multirow{2}{*}{$\calL_{\mathrm{MSE}}$} & \multirow{2}{*}{$\calL_{\mathrm{KL}}$} & \multirow{2}{*}{$\calL_{\mathrm{DP}}$} & \multirow{2}{*}{$\calL_{\mathrm{margin}}$} & {ResNet-18}  \tabularnewline
     & &  &  &&  & W2A2  \tabularnewline
     \hline
     Genie-M~\cite{Genie} & &  &  &&  & 53.71 \tabularnewline
    \hline 
    (a) & \checkmark &  &  & &   & 53.45\tabularnewline
    (b) & \checkmark & \checkmark &  & & &53.64 \tabularnewline
    (c) & \checkmark &  & \checkmark & & & 53.89\tabularnewline
    (d) & \checkmark &  & &\checkmark& & 54.09 \tabularnewline
    (e) & \checkmark &  &  & \checkmark&  \checkmark & \textbf{54.22} \tabularnewline
    \hline 
    \end{tabular}
\end{table}
\subsection{Ablation studies}
\label{subsec:ablation}
\paragraph{\textbf{Comparitions of information preservation losses.}}
We conduct ablation studies to compare the three different information preservation losses \cref{eq:MSE}, \cref{eq:KL}, and \cref{eq:PKT}. 
As shown in (\ref{eq:loss_T_alpha_finalLP}), the final loss for updating the transformation network $T$
is a combination of three different losses, consisting of the validation loss $\mathcal{L}_{val}$, identity prevention loss $\mathcal{L}_{margin}$, and one of the three information preservation losses.  We conduct experiments on the ResNet-18 model for the 2/2 bit-width setting. The results are presented in Table~\ref{tab:loss_combined}. The results show that the classification accuracy decreases compared to the Genie baseline~\cite{Genie} when using only $\calL_{val}$ (setting (a)). Meanwhile, combining $\calL_{val}$ with $\mathcal{L}_{DP}$ (setting (d)) results in improvements of 0.45\% and 0.2\% compared to the combinations of $\calL_{val}$ with $\calL_{MSE}$ (setting (b)), and $\calL_{val}$ with $\calL_{KL}$ (setting (c)), respectively. 
Furthermore, using additional $\mathcal{L}_{margin}$ (settings (e)) results in even further improvements. This shows that both $\mathcal{L}_{margin}$ and $\mathcal{L}_{DP}$ are essential for the final results.
For the remaining results in the following sections, $\mathcal{L}_{DP}$ is used in the $\mathcal{L}_{T}$ (\cref{eq:loss_T_alpha_finalLP}). \textcolor{black}{The ablation studies of the hyper-parameters are provided in the supplementary material}.

\begin{table}[h!]
    \centering
    \caption{Comparisons of Top-1 classification accuracy (\%) with the state of the art on ImageNet dataset. The notation $*$ indicates that the input (activation) of the second layer is maintained at 8-bit precision following BRECQ \cite{BRECQ} setting. The result denoted with $\ddagger$ is
    reproduced using the official released code of the corresponding paper.
    }
    \label{tab:sota}
    \begin{tabular}{l c c c c}
    \toprule
    \multirow{2}{4em}{\bfseries Method} & \bfseries Bit-width & \bfseries ResNet-18 & \bfseries ResNet-50 & \bfseries MobileNetV2 \\ 
     &  (W/A) &   &  &   \\
    \midrule
   FP & 32/32 &  71.01 &  {76.63} &  {72.20} \\
    \midrule
    AdaRound \cite{AdaRound} & \multirow{9}{*}{4/4} & {67.96} & {73.88} & {61.52} \\
    BRECQ* \cite{BRECQ} & & 69.60 & 75.05 & 66.57 \\
    QDrop \cite{Wei2022QDropRD} & & 69.10 & 75.03 & 67.89 \\
    QDrop* \cite{Wei2022QDropRD} & & 69.62 & 75.45 & 68.84 \\
    PD-Quant \cite{liu2023pd} & & 69.23 & 75.16 & 68.19 \\
    
    Genie-M \cite{Genie} & & 69.35 & {75.21} & {68.65} \\

    Bit-Shrinking* \cite{lin2023bit} &  & {69.94} & \textbf{76.04} & {69.02} \\
    
    \rowcolor{Gray!25} MetaAug (Ours) & & 69.48 & 75.29 & 68.76 \\
    \rowcolor{Gray!25} MetaAug* (Ours) &  & \textbf{69.97}  & 75.78  & \textbf{69.22} \\

    \midrule
    AdaRound \cite{AdaRound} & \multirow{8}{*}{3/3} & 64.14 & 68.40 & 41.52 \\
    BRECQ* \cite{BRECQ} &  & 64.80 & 70.29 & 53.34 \\
    QDrop \cite{Wei2022QDropRD} &  & 65.56 & 71.07 & 54.27 \\
    QDrop* \cite{Wei2022QDropRD} &  & 66.75 & 72.38 & 57.98 \\
    Genie \cite{Genie} &  & 66.16 & 71.61 & 57.54 \\
    Bit-Shrinking* \cite{lin2023bit} &  & {67.12} & 72.91 & {58.66} \\
    \rowcolor{Gray!25} MetaAug (Ours) &  & 66.37 & 71.73 & 57.77 \\
    \rowcolor{Gray!25} MetaAug* (Ours) &  & \textbf{67.66} & \textbf{73.04}  & \textbf{59.87} \\

    \midrule
    
    BRECQ* \cite{BRECQ} & \multirow{8}{*}{2/4} & 64.80 & 70.29 & 53.34 \\
    QDrop \cite{Wei2022QDropRD} &  & 64.66 & 70.08 & 52.92 \\
    QDrop* \cite{Wei2022QDropRD} &  & 65.25 & 70.65 & 54.22 \\
  
    PD-Quant \cite{liu2023pd} &  & 65.17 & 70.77 & 55.17 \\
    Genie-M \cite{Genie} &  & {65.77} & 70.51 & 56.38 \\
    
    Bit-Shrinking* \cite{lin2023bit} &  & 65.77 & 71.11 & 54.88 \\
    
    \rowcolor{Gray!25} MetaAug (Ours) &  & 66.01 & 70.76 & 56.45 \\
    \rowcolor{Gray!25} MetaAug* (Ours) &  & \textbf{66.48} & \textbf{71.48} & \textbf{56.65} \\
    \midrule

    BRECQ* \cite{BRECQ} & \multirow{8}{*}{2/2} & 42.54 & 29.01 & 0.24 \\
    QDrop \cite{Wei2022QDropRD} &  & 51.14 & 54.74 & 8.46 \\
    
    QDrop* \cite{Wei2022QDropRD} &  & 54.72 & 58.67 & 13.05 \\
    PD-Quant \cite{liu2023pd} &  & 53.14 & 57.16 & 13.76 \\
    Genie-M \cite{Genie} &  & 53.71 & 56.71 & $16.25^\ddagger$ \\

    Bit-Shrinking* \cite{lin2023bit} &  & 57.33 & 59.03 & {18.23} \\
    \rowcolor{Gray!25} MetaAug (Ours) &  & 54.22 & 57.30 &  16.97\\
    \rowcolor{Gray!25} MetaAug* (Ours) &  & \textbf{57.89} & \textbf{60.50} & \textbf{19.61} \\
    \bottomrule
    
    \end{tabular}
    \end{table}
\subsection{Comparisons with the state of the art}
In this section, we compare our proposed method against the state-of-the-art methods for PTQ, including AdaRound~\cite{AdaRound}, BRECQ~\cite{BRECQ}, QDrop~\cite{Wei2022QDropRD}, PD-Quant~\cite{liu2023pd}, Genie-M~\cite{Genie}, and Bit-Shrinking~\cite{lin2023bit}. The results of competitors are cited from the corresponding papers except for the 2/2 setting of Genie-M~\cite{Genie} with the MobileNetV2 network which is reproduced by their official release code. \Cref{tab:sota} presents the comparative results of our proposed MetaAug and other state-of-the-art approaches when evaluating on the ImageNet dataset. It is clear that our proposed method outperforms the other methods across various network architectures. Compared to the current state-of-the-art approaches, Genie-M~\cite{Genie}, our proposed method consistently outperforms Genie-M in all bit-width settings. The improvement is clearer in the 2/2 settings, with an improvement of 0.51\%, 0.59\%, and 0.72\% for ResNet-18, ResNet-50, and MobileNetV2, respectively. When activation of the second layer is kept at 8-bit precision, following BRECQ~\cite{BRECQ} setting, our proposed method outperforms the current state-of-the-art, Bit-Shrinking~\cite{lin2023bit} in all bit-width settings except for the ResNet-50 in 4/4 setting. The improvement is clearer with the highest improvement over Bit-shrinking being 1.47\% for ResNet-50 in the 2/2 setting, which confirms the effectiveness of our proposed approach.
\begin{table}[h!]
    \caption{Top-1 classification accuracy (\%) of ResNet18 on 1024 calibration images (train set), and testing images of the ImageNet dataset, and the gap between accuracy on the calibration set and the test set.}
    \label{tab:overfitting}
    \centering
    \resizebox{1.0\columnwidth}{!}{
  \begin{tabular}{l|c|c|c|c}
  \hline
  \multirow{2}{4em}{ \bfseries Method} & {\bfseries Bit-width} & {\bfseries Accuracy on} & {\bfseries Accuracy on the} & \bfseries Train-test \\ 
  &  \bfseries (W/A) & \bfseries the test set & \bfseries  calibration set &\bfseries accuracy gap\\
  \hline
   FP &  32/32 &  71.01 &  85.16 &  14.15 \\
  \hline
    QDrop~\cite{Wei2022QDropRD} &\multirow{4}{*}{2/2} & 51.14 & 77.53 & 26.39\\
    PD-Quant~\cite{liu2023pd}  &  & 53.14  & 83.30 &30.16 \\
    Genie-M~\cite{Genie} &  & 53.77 & 80.18  & 27.01\\
    MetaAug (Ours) & & 54.22 & 77.64 & \textbf{23.42}\\
  \hline
   QDrop~\cite{Wei2022QDropRD}& \multirow{4}{*}{2/4} & 64.66 & 81.64  & 16.98\\
 PD-Quant~\cite{liu2023pd}  &  & 65.17  & 84.38& 19.21\\
  Genie-M~\cite{Genie} &  & 65.77 & 84.18 & 18.41\\
  MetaAug (Ours) & & 66.01 & 82.91 & \textbf{16.90} \\
  \hline
  \end{tabular}
  }
  \end{table}
\begin{figure}[h!]
        \centering
         \includegraphics[width=1\textwidth]{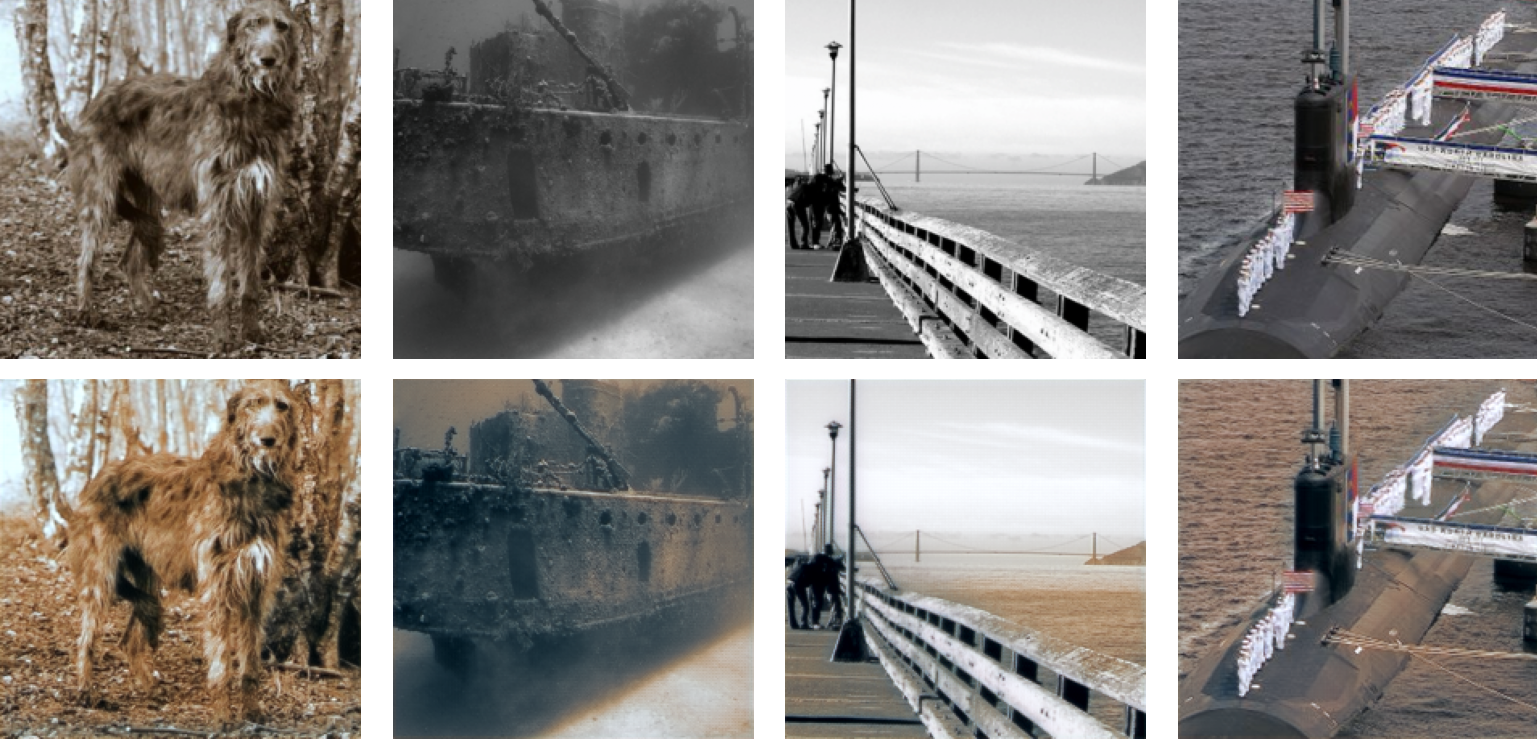}
        \caption{Visualization of the original calibration images (the first row) and the corresponding modified images (the second row) produced by the transformation network. 
        }
    \label{fig:generated_images}
\end{figure}
\paragraph{\textbf{Mitigating overfitting.}}
To investigate the benefits of our approach in addressing the overfitting problem, we conduct experiments demonstrating the performance of our methods over the calibration set (i.e., the train set) and the test set, compared to other state-of-the-art methods, including PD-Quant\cite{liu2023pd}, QDrop\cite{Wei2022QDropRD}, and Genie\cite{Genie}. The results are presented in Table \ref{tab:overfitting}. It is clear that our proposed method not only achieves the highest accuracy on the test set compared to other models but also yields the smallest train-test accuracy gap. 
Compared to QDrop~\cite{Wei2022QDropRD}, while there is a marginal difference between our proposed and QDrop~\cite{Wei2022QDropRD} 
in terms of the train-test accuracy gap (16.98\% versus 16.90\%), our approach achieves significant improvements of  3.08\% and 1.35\% over QDrop~\cite{Wei2022QDropRD} in the test set for the 2/2 and 2/4 settings, respectively.
\paragraph{\textbf{Visualization.}}
Some original calibration images and the corresponding images produced by the transformation network are presented in \cref{fig:generated_images}. 
The images are produced with 
the setting 2/2 with the ResNet18 model. 
As shown in \cref{fig:generated_images}, the modified images change the appearance while still preserving the semantic information of the original calibration images.
\begin{table}[h!]
    \caption{Comparative Top-1 classification accuracy (\%) with the 2/2 setting with ResNet-18 between our method and other augmentation approaches.}
    \label{tab:abl_augmentation}
    \centering
  \begin{tabular}{l|c|c}
  \hline
  \multirow{2}{4em}{ \bfseries Augmentation} & {\bfseries Bit-width} & {\bfseries ResNet-18} \\ 
  &   (W/A) & (FP: 71.01) \\
  \hline
  \hline
   Genie-M (no augmentation)~\cite{Genie} & \multirow{11}{*}{2/2} & 53.71\\
   Contrast &  & 53.57 \\
    Brightness &  & 53.35   \\
    Random Flip &  &  53.93  \\
    Random Rotation &  &  53.95 \\
   Random flip + Rotation + Brightness &  & 53.87 \\
   Mixup~\cite{mixup} &  & 54.05  \\
   Cutmix~\cite{cutmix}  &  & 54.15  \\
   MetaAug (Ours) &  & 54.22  \\ 
   MetaAug (Ours) + Mixup &  & 54.35  \\
   MetaAug (Ours) + Cutmix &  &  \textbf{54.63} \\
  \hline
  \end{tabular}
  \end{table}
\subsection{Additional results}
\paragraph{\textbf{Comparisons with other augmentation approaches.}} We compare the results of our proposed method with various augmentation strategies, including traditional photometric data augmentation, such as contrast and brightness adjustments, and geometric data augmentation, such as random flipping and random rotation. We also investigate advanced augmentation methods, i.e.,  Mixup~\cite{mixup} and Cutmix~\cite{cutmix}. The results of these augmentation techniques are presented in \Cref{tab:abl_augmentation}. The results show that the considered 
geometric augmentation strategies improve the performance over the baseline Genie-M~\cite{Genie}, while an opposite observation is with the considered photometric. 
The results also show that our proposed method outperforms all compared augmentation strategies, including the advanced augmentation methods Mixup~\cite{mixup} and Cutmix~\cite{cutmix}. This confirms the effectiveness of the proposed method.  
Furthermore, combining Mixup~\cite{mixup} or Cutmix~\cite{cutmix} augmentation with our proposed method yields even more improvements. 
This indicates that our approach and existing advanced augmentation techniques can complement each other when used together.

\section{Conclusion}
\label{sec:conclusion}
In this paper, we propose a novel meta-learning based approach to mitigate the overfitting problem in post-training quantization. Specifically, we jointly optimize a transformation network, which is used to modify the original calibration data, and a quantized model in a bi-level optimization process. Additionally, we explore different losses, including an advanced distribution preservation loss, and propose using a margin loss for training transformation network so that the outputs of the network preserve the feature information of the original calibration data while preventing it from becoming an identity mapping. We extensively evaluate our proposed approach on the ImageNet dataset across various network architectures, demonstrating that the proposed method outperforms current state-of-the-art PTQ methods. A limitation of the current work is that the transformation network does not perform geometric transformations. In future work, we can consider designing a transformation network that also encodes geometric transformations, e.g., by integrating the Spatial Transformer module~\cite{jaderberg2015spatial} to spatially transform regions of images. This will result in more diverse augmented images, which could improve the effectiveness of the proposed approach.

\section*{Acknowledgements}
Trung Le and Dinh Phung were supported by ARC DP23 grant DP230101176 and by the Air Force Office of Scientific Research under award number FA2386-23-1-4044.

%
%
\bibliographystyle{splncs04}
\bibliography{main}

\begin{thebibliography}{10}
\providecommand{\url}[1]{\texttt{#1}}
\providecommand{\urlprefix}{URL }
\providecommand{\doi}[1]{https://doi.org/#1}

\bibitem{sharp-maml}
Abbas, M., Xiao, Q.W., Chen, L., Chen, P.Y., Chen, T.: Sharp-{MAML}:
  Sharpness-aware model-agnostic meta learning. In: ICML (2022)

\bibitem{andrychowicz2016learning}
Andrychowicz, M., Denil, M., Gomez, S., Hoffman, M.W., Pfau, D., Schaul, T.,
  Shillingford, B., De~Freitas, N.: Learning to learn by gradient descent by
  gradient descent. In: NIPS. vol.~29 (2016)

\bibitem{cai2020zeroq}
Cai, Y., Yao, Z., Dong, Z., Gholami, A., Mahoney, M.W., Keutzer, K.: Zeroq: A
  novel zero shot quantization framework. In: CVPR (2020)

\bibitem{chen2019metaquant}
Chen, S., Wang, W., Pan, S.J.: Metaquant: Learning to quantize by learning to
  penetrate non-differentiable quantization. NeurIPS  (2019)

\bibitem{BinaryConnect}
Courbariaux, M., Bengio, Y., David, J.P.: Binaryconnect: Training deep neural
  networks with binary weights during propagations. In: NIPS. pp. 3123--3131
  (2015)

\bibitem{cubuk2020randaugment}
Cubuk, E.D., Zoph, B., Shlens, J., Le, Q.V.: Randaugment: Practical automated
  data augmentation with a reduced search space. In: CVPR. pp. 702--703 (2020)

\bibitem{defossez2021differentiable}
D{\'e}fossez, A., Adi, Y., Synnaeve, G.: Differentiable model compression via
  pseudo quantization noise. TMLR  (2022)

\bibitem{Duan2016RL2FR}
Duan, Y., Schulman, J., Chen, X., Bartlett, P.L., Sutskever, I., Abbeel, P.:
  Rl: Fast reinforcement learning via slow reinforcement learning. ArXiv
  (2016)

\bibitem{LSQ}
Esser, S.K., McKinstry, J.L., Bablani, D., Appuswamy, R., Modha, D.S.: {Learned
  Step Size Quantization}. In: ICLR (2020)

\bibitem{sign-maml}
Fan, C., Ram, P., Liu, S.: Sign-{MAML}: Efficient model-agnostic meta-learning
  by {SignSGD}. ArXiv  (2021)

\bibitem{finn2017model}
Finn, C., Abbeel, P., Levine, S.: Model-agnostic meta-learning for fast
  adaptation of deep networks. In: ICML (2017)

\bibitem{DSQ}
Gong, R., Liu, X., Jiang, S., Li, T., Hu, P., Lin, J., Yu, F., Yan, J.:
  Differentiable soft quantization: Bridging full-precision and low-bit neural
  networks. In: ICCV (2019)

\bibitem{han2015deep}
Han, S., Mao, H., Dally, W.J.: Deep {C}ompression: Compressing deep neural
  network with pruning, trained quantization and huffman coding. In: ICLR
  (2016)

\bibitem{resnet}
He, K., Zhang, X., Ren, S., Sun, J.: Deep residual learning for image
  recognition. In: CVPR (2016)

\bibitem{jaderberg2015spatial}
Jaderberg, M., Simonyan, K., Zisserman, A., Kavukcuoglu, K.: Spatial
  transformer networks. NeurIPS  (2015)

\bibitem{Genie}
Jeon, Y., Lee, C., Kim, H.y.: Genie: Show me the data for quantization. In:
  CVPR (2023)

\bibitem{mix-maml}
Jia, J., Feng, X., Yu, H.: Few-shot classification via efficient meta-learning
  with hybrid optimization. Engineering Applications of Artificial Intelligence
   (2024)

\bibitem{Kim2023MetaMixMP}
Kim, H.B., Lee, J.H., Yoo, S., Kim, H.S.: {MetaMix}: Meta-state precision
  searcher for mixed-precision activation quantization. In: AAAI (2024)

\bibitem{Kingma2014AdamAM}
Kingma, D.P., Ba, J.: Adam: A method for stochastic optimization. In: ICLR
  (2015)

\bibitem{Koch2015SiameseNN}
Koch, G., Zemel, R., Salakhutdinov, R.: Siamese neural networks for one-shot
  image recognition. In: ICML deep learning workshop (2015)

\bibitem{BRECQ}
Li, Y., Gong, R., Tan, X., Yang, Y., Hu, P., Zhang, Q., Yu, F., Wang, W., Gu,
  S.: {{BRECQ}}: Pushing the limit of post-training quantization by block
  reconstruction. In: ICLR (2021)

\bibitem{li2017meta}
Li, Z., Zhou, F., Chen, F., Li, H.: Meta-sgd: Learning to learn quickly for
  few-shot learning. arXiv preprint arXiv:1707.09835  (2017)

\bibitem{lin2023bit}
Lin, C., Peng, B., Li, Z., Tan, W., Ren, Y., Xiao, J., Pu, S.: Bit-{S}hrinking:
  Limiting instantaneous sharpness for improving post-training quantization.
  In: CVPR (2023)

\bibitem{liu2023pd}
Liu, J., Niu, L., Yuan, Z., Yang, D., Wang, X., Liu, W.: Pd-quant:
  Post-training quantization based on prediction difference metric. In: CVPR
  (2023)

\bibitem{ma2023solving}
Ma, Y., Li, H., Zheng, X., Xiao, X., Wang, R., Wen, S., Pan, X., Chao, F., Ji,
  R.: Solving oscillation problem in post-training quantization through a
  theoretical perspective. In: CVPR (2023)

\bibitem{tsne}
Van~der Maaten, L., Hinton, G.: Visualizing data using t-{SNE}. Journal of
  machine learning research  (2008)

\bibitem{muller2021trivialaugment}
M{\"u}ller, S.G., Hutter, F.: Trivialaugment: Tuning-free yet state-of-the-art
  data augmentation. In: ICCV (2021)

\bibitem{AdaRound}
Nagel, M., Amjad, R.A., Van~Baalen, M., Louizos, C., Blankevoort, T.: Up or
  down? adaptive rounding for post-training quantization. In: ICML (2020)

\bibitem{nagel2019data}
Nagel, M., Baalen, M.v., Blankevoort, T., Welling, M.: Data-free quantization
  through weight equalization and bias correction. In: CVPR (2019)

\bibitem{nahshan2021loss}
Nahshan, Y., Chmiel, B., Baskin, C., Zheltonozhskii, E., Banner, R., Bronstein,
  A.M., Mendelson, A.: Loss aware post-training quantization. Machine Learning
  \textbf{110}(11-12),  3245--3262 (2021)

\bibitem{fo-maml}
Nichol, A., Achiam, J., Schulman, J.: On first-order meta-learning algorithms.
  ArXiv  (2018)

\bibitem{PKT}
Passalis, N., Tefas, A.: Learning deep representations with probabilistic
  knowledge transfer. In: ECCV (2018)

\bibitem{XNOR-Net}
Rastegari, M., Ordonez, V., Redmon, J., Farhadi, A.: {XNOR-Net}: Imagenet
  classification using binary convolutional neural networks. In: ECCV (2016)

\bibitem{ravi2017optimization}
Ravi, S., Larochelle, H.: Optimization as a model for few-shot learning. In:
  ICLR (2017)

\bibitem{Unet}
Ronneberger, O., Fischer, P., Brox, T.: U-{N}et: Convolutional networks for
  biomedical image segmentation. In: MICCAI (2015)

\bibitem{imagenet}
Russakovsky, O., Deng, J., Su, H., Krause, J., Satheesh, S., Ma, S., Huang, Z.,
  Karpathy, A., Khosla, A., Bernstein, M., Berg, A., Li, F.F.: Image{N}et large
  scale visual recognition challenge. IJCV  (2015)

\bibitem{Mobilenetv2}
Sandler, M., Howard, A., Zhu, M., Zhmoginov, A., Chen, L.C.: Mobile{N}et{V}2:
  Inverted residuals and linear bottlenecks. In: CVPR (2018)

\bibitem{mann}
Santoro, A., Bartunov, S., Botvinick, M., Wierstra, D., Lillicrap, T.:
  Meta-learning with memory-augmented neural networks. In: ICML (2016)

\bibitem{Satorras2017FewShotLW}
Satorras, V.G., Bruna, J.: Few-shot learning with graph neural networks. In:
  ICLR (2018)

\bibitem{shin2023nipq}
Shin, J., So, J., Park, S., Kang, S., Yoo, S., Park, E.: Nipq: Noise
  proxy-based integrated pseudo-quantization. In: CVPR (2023)

\bibitem{Snell2017PrototypicalNF}
Snell, J., Swersky, K., Zemel, R.S.: Prototypical networks for few-shot
  learning. In: NeurIPS (2017)

\bibitem{Sung2017LearningTC}
Sung, F., Yang, Y., Zhang, L., Xiang, T., Torr, P.H.S., Hospedales, T.M.:
  Learning to compare: Relation network for few-shot learning. CVPR  (2017)

\bibitem{Vinyals2016MatchingNF}
Vinyals, O., Blundell, C., Lillicrap, T.P., Kavukcuoglu, K., Wierstra, D.:
  Matching networks for one shot learning. In: NeurIPS (2016)

\bibitem{Wang2016LearningTR}
Wang, J.X., Kurth-Nelson, Z., Soyer, H., Leibo, J.Z., Tirumala, D., Munos, R.,
  Blundell, C., Kumaran, D., Botvinick, M.M.: Learning to reinforcement learn.
  ArXiv  (2016)

\bibitem{Wang2020AutomaticLH}
Wang, T., Wang, J., Xu, C., Xue, C.: Automatic low-bit hybrid quantization of
  neural networks through meta learning. ArXiv  (2020)

\bibitem{Wei2022QDropRD}
Wei, X., Gong, R., Li, Y., Liu, X., Yu, F.: {QD}rop: Randomly dropping
  quantization for extremely low-bit post-training quantization. In: ICLR
  (2022)

\bibitem{xu2020generative}
Xu, S., Li, H., Zhuang, B., Liu, J., Cao, J., Liang, C., Tan, M.: Generative
  low-bitwidth data free quantization. In: ECCV (2020)

\bibitem{yang2019quantization}
Yang, J., Shen, X., Xing, J., Tian, X., Li, H., Deng, B., Huang, J., Hua, X.S.:
  Quantization networks. In: CVPR (2019)

\bibitem{Youn2022BitwidthAdaptiveQN}
Youn, J., Song, J., Kim, H.S., Bahk, S.: Bitwidth-adaptive quantization-aware
  neural network training: A meta-learning approach. In: ECCV (2022)

\bibitem{cutmix}
Yun, S., Han, D., Oh, S.J., Chun, S., Choe, J., Yoo, Y.J.: Cut{M}ix:
  Regularization strategy to train strong classifiers with localizable
  features. ICCV  (2019)

\bibitem{mixup}
Zhang, H., Ciss{\'e}, M., Dauphin, Y., Lopez-Paz, D.: mixup: Beyond empirical
  risk minimization. In: ICLR (2018)

\bibitem{zheng2022leveraging}
Zheng, D., Liu, Y., Li, L.: Leveraging inter-layer dependency for post-training
  quantization. NeurIPS  (2022)

\end{thebibliography}

\clearpage
\setcounter{table}{0}
\setcounter{section}{0}
\setcounter{figure}{0}
\renewcommand{\thetable}{A.\arabic{table}}
\renewcommand{\thesection}{\textbf{A.\arabic{section}}}
\renewcommand{\thefigure}{A.\arabic{figure}}
\section*{\centering{Supplementary Materials}}

\section{Hyper-parameter settings}

\paragraph{\textbf{Hyper-parameters $\lambda_1$, $\lambda_2$, $\lambda_3$.}}

Regarding hyper-parameters $\lambda_1$, $\lambda_2$, and $\lambda_3$ in Eq. (15) in the main paper,  these parameters control the impacts of validation loss, margin loss, and preservation loss on the overall loss for learning the transformation network $T$. We present ablation studies on the choice of hyperparameters $\lambda_1$, $\lambda_2$, and $\lambda_3$ on the ImageNet dataset. For all experiments in this supplementary material, 
$\mathcal{L}_{DP}$ is used in the $\mathcal{L}_{T}$ (Eq. (15) in the main paper). 
\noindent \textbf{For ablation studies for parameter $\lambda_1$}, we vary the value of $\lambda_1$ from 1 to 10 and fix the value of $\lambda_2=0$, and $\lambda_3=3\times 10^4$ . The results are shown in \Cref{tab:param_lambda1}. The results show that $\lambda_1$'s range from 5 to 10 often leads to better performance for the 2/2 and 2/4 settings, and the proposed method does not show high sensitivity to the choice of $\lambda_1$.

\begin{table}[h]
    \caption{Ablation study for hyper-parameter $\lambda_1$ of validation loss in Eq. (15). The results are on the ImageNet dataset with 2/2 and 2/4 settings.}
    \label{tab:param_lambda1}
    \centering
    \begin{tabular}{l|ccccccc}
      \hline 
       $\lambda_1$ & 1 & 2 &3 &5 &8&10  \tabularnewline
      \hline 
        2/2 & 54.06 & 54.03 & 53.99 & \textbf{54.09} & 54.01 & 54.05   \\
      2/4 & 65.88 & 65.94 & 65.91 & 65.96 & 65.93  & \textbf{66.03} \\
      \bottomrule
    \end{tabular}
  \end{table}

  \noindent\textbf{For ablation studies for parameter $\lambda_2$}, we vary the value of $\lambda_2$ from 0.1 to 1, and fix the value of  $\lambda_1=5$, and $\lambda_3=3\times 10^4$. The $\epsilon$ in Eq. (14) is set to 0.1. The results are shown in Table~\ref{tab:param_lambda2}. The results indicate that $\lambda_2$'s range from 0.2 to 0.5 yields better performance.

  \begin{table}[h]
    \caption{Ablation study for hyper-parameter $\lambda_2$ of margin loss in Eq. (15). The results are on the ImageNet dataset with 2/2 and 2/4 settings.}
    \label{tab:param_lambda2}
    \centering
    \begin{tabular}{l|cccccc}
      \hline 
       $\lambda_2$ & $0.1$ & 0.2 & 0.3 & 0.5 & 0.8 & 1.0   \tabularnewline
       \hline
        2/2 & 54.02 & 53.90 & 54.17 & \textbf{54.22} & 54.15 & 54.19  \\
      2/4 & 65.90 & \textbf{66.05} & 65.80 & 66.01 & 65.83 & 65.97   \\
      \bottomrule
    \end{tabular}
  \end{table}

  \noindent\textbf{For ablation studies for parameter $\lambda_3$}, we vary the value of $\lambda_3$ from $10^4$ to $10^5$, and fix the value of $\lambda_1=5$, and $\lambda_2=0.5$. The $\epsilon$ in Eq. (14) is set to 0.1. The results are shown in Table~\ref{tab:param_lambda3}. The results show that the $\lambda_3$'s range from $2\times 10^4$ to $5\times 10^4$ often leads to higher performance, while the performance may not be sensitive to the choice of $\lambda_3$.

 \begin{table}[h]
    \caption{Ablation study for hyper-parameter $\lambda_3$ of distribution preservation loss in Eq. (15). The results are on the ImageNet dataset with 2/2 and 2/4 settings.}
    \label{tab:param_lambda3}
    \centering
    \begin{tabular}{l|cccccccc}
      \hline 
       $\lambda_3$ & $1\times 10^4$ & $2\times10^4$ & $3\times10^4$ & $5\times10^4$ &  $8\times10^4$ & $10\times10^4$ & \tabularnewline
      \hline 
        2/2 & 54.10 & 54.04 & \textbf{54.22} & 54.12 &54.08 & 54.15  \\
      2/4 & 65.87 & \textbf{66.05} & {66.01} & 66.02 & 65.81 & 65.74 \\
      \bottomrule
    \end{tabular}
  \end{table}
\paragraph{\textbf{The sensitivity of hyper-parameter $\epsilon$ in Eq. (14).}} 
We conduct ablation study for the sensitivity of hyper-parameter $\epsilon$. We vary the value of $\epsilon$ from 0.1 to 2 and fix the value of $\lambda_1=5$, $\lambda_2=0.5$, and $\lambda_3=3\times10^4$. The results are presented in~\Cref{tab:epsilon}. The results show that the best value of $\epsilon$ is 0.3 for the 2/2 setting and 0.1 for the 2/4 setting. Setting $\epsilon$ higher (e.g., $\epsilon=2$) results in modified images that could not retain the intrinsic information from the original images. 

 \begin{table}[h]
    \caption{Ablation study for hyper-parameter $\epsilon$ of margin loss in Eq. (15). The results are on the ImageNet dataset with 2/2 and 2/4 settings.}
    \label{tab:epsilon}
    \centering
    \begin{tabular}{l|cccccccc}
      \hline 
       $\epsilon$ & 0.1 & 0.2 & 0.3 & 0.5 & 0.8 & 1 & 2 \tabularnewline
      \hline 
        2/2 & 54.22 & 54.03& \textbf{54.44}& 53.86& 54.02 & 54.10 & 53.90  \\
        2/4 & \textbf{66.01} & 66.95 & 65.91 & 65.87 & 65.86 & 65.75 & 65.61 \\
      \bottomrule
    \end{tabular}
  \end{table}
\section{Additional comparisons with automated data augmentation}

In addition to traditional augmentation techniques (e.g. Random Flip, Rotation, Brightness) and advanced augmentation methods (e.g. MixUp, CutMix) that have been presented in the main paper, we also compare the results of MetaAug with automated data augmentation approaches
including 
RandAugment~\cite{cubuk2020randaugment}, and TrivialAugment~\cite{muller2021trivialaugment}. 
These augmentations are combinations of multiple transforms, either geometric or photometric, or both.
Following ~\cite{cubuk2020randaugment,muller2021trivialaugment}, we adopt the 14 different transformations: \textit{identity, autocontrast, equalize, posterize, rotate, solarize, shear-x, shear-y, translate-x, translate-y, color, contrast, brightness, and sharpness}. Among those transformations, the photometric transformations include: \textit{autocontrast, equalize, posterize, solarize, color, contrast, brightness, and sharpness}. Meanwhile, the geometric transformations include: \textit{rotate, shear-x, shear-y, translate-x, and translate-y}. 

\begin{table}[t!]
    \caption{Comparative Top-1 classification accuracy (\%) on the ImageNet dataset with the 2/2 setting with ResNet-18 between our proposed method and  automated data augmentation.
    }
    \label{tab:abl_original_augment_learning}
    \centering
  \begin{tabular}{l|c|c}
  \hline
  \multirow{2}{4em}{ \bfseries Augmentation} & {\bfseries Bit-width} & {\bfseries ResNet-18} \\ 
  &   (W/A) & (FP: 71.01) \\
  \hline
  \hline
   Genie-M (no augmentation)~\cite{Genie} & \multirow{6}{*}{2/2} & 53.71\\
   TrivialAugment~\cite{muller2021trivialaugment} &  & 53.86 \\
    RandAugment~\cite{cubuk2020randaugment} &  & 53.55  \\
   MetaAug (Ours) &  &  \textbf{54.22} \\ 
   MetaAug (Ours) + TrivialAugment &  & 54.06  \\
   MetaAug (Ours) + RandAugment &  &  53.92 \\
  \hline
  \end{tabular}
  \end{table}
\paragraph{\textbf{Automated data augmentation.}} We first compare the proposed MetaAug with automated data augmentation approaches using 14 transformations that include both photometric and geometric transformations. The results presented in~\Cref{tab:abl_original_augment_learning} show that TrivalAugment and RandAugment seem not to impact the original Genie-M~\cite{Genie}, and the performance is even decreased 
with RandAugment. Additionally, the combination of images produced by those methods and images produced by our transformation network also leads to performance decreases. 
\begin{table}[h!]
    \caption{Comparative Top-1 classification accuracy (\%) on the ImageNet dataset with the 2/2 setting with ResNet-18 between our proposed method and automated photometric data augmentation.
    }
\label{tab:abl_photometric_augment_learning}
    \centering
  \begin{tabular}{l|c|c}
  \hline
  \multirow{2}{4em}{ \bfseries Augmentation} & {\bfseries Bit-width} & {\bfseries ResNet-18} \\ 
  &   (W/A) & (FP: 71.01) \\
  \hline
  \hline
   Genie-M (no augmentation)~\cite{Genie} & \multirow{6}{*}{2/2} & 53.71\\
   TrivialAugment (photometric)~\cite{muller2021trivialaugment} &  &  53.53\\
   RandAugment (photometric)~\cite{cubuk2020randaugment} &  & 53.46 \\
   MetaAug (Ours) &  & \textbf{54.22}  \\ 
   MetaAug (Ours) + TrivialAugment (photometric) &  & 53.89 \\
   MetaAug (Ours) + RandAugment (photometric) &  &  53.80 \\
  \hline
  \end{tabular}
  \end{table}
\paragraph{\textbf{Automated photometric data augmentation.}} ~\Cref{tab:abl_photometric_augment_learning} shows the results when automated data augmentation only contains the photometric transformations. The results indicate that the combination of images produced by automated photometric data augmentation and images produced by our transformation network results in performance decreases. In addition, automated photometric data augmentation methods result in performance decreases of 0.18\% and 0.25\% over baseline Genie-M~\cite{Genie} for the TrivialAugment~\cite{muller2021trivialaugment} and RandAugment~\cite{cubuk2020randaugment} settings, respectively. This indicates that simple photometric augmentation could potentially reduce the performance of PTQ. 
\begin{table}[t!]
    \caption{Comparative Top-1 classification accuracy (\%) on ImageNet dataset with the 2/2 setting with ResNet-18 between our proposed method and automated geometric data augmentation.
    }
    \label{tab:abl_geometric_augment_learning}
    \centering
  \begin{tabular}{l|c|c}
  \hline
  \multirow{2}{4em}{ \bfseries Augmentation} & {\bfseries Bit-width} & {\bfseries ResNet-18} \\ 
  &   (W/A) & (FP: 71.01) \\
  \hline
  \hline
   Genie-M (no augmentation)~\cite{Genie} & \multirow{6}{*}{2/2} & 53.71\\
   TrivialAugment (Geometric)~\cite{muller2021trivialaugment} &  & 54.04  \\
   RandAugment (Geometric)~\cite{cubuk2020randaugment} &  & 54.06 \\
   MetaAug (Ours) &  & 54.22 \\ 
   MetaAug (Ours) + TrivialAugment (Geometric)&  & \textbf{54.52}  \\
   MetaAug (Ours) + RandAugment (Geometric) &  &  54.41 \\
  \hline
  \end{tabular}
  \end{table}

\paragraph{\textbf{Automated geometric data augmentation.}} ~\Cref{tab:abl_geometric_augment_learning} shows the result when automated data augmentation contains only the combination of the geometric transformations. 
The results show that these augmentation techniques can enhance the performance of PTQ. Specifically, using automated geometric data augmentation achieves improvements over the baseline Genie-M~\cite{Genie} by 0.33\% and 0.35\% for TrivialAugment and RandAugment, respectively, in the 2/2 setting. 
Combining the images produced by our MetaAug with images produced by automated geometric augmentation, as shown in~\Cref{tab:abl_geometric_augment_learning}, leads to a significant enhancement in PTQ performance, achieving the highest results in this table. The improvements over the baseline Genie-M (no augmentation)~\cite{Genie} are 0.81\% and 0.70\% for TrivialAugment and RandAugment, respectively, in the 2/2 setting. Meanwhile, the improvements over MetaAug alone are 0.3\% and 0.19\% for TrivialAugment and RandAument, respectively. This indicates that our approach MetaAug and automated geometric data augmentation can complement each other when used together.

\begin{table}[h!]
\label{tab:performance_gain_random}
  \centering
   \caption{The comparative performance of PTQ with various calibration data sizes on ResNet-18 in the 2/2 setting.
}
  \begin{tabular}{l|ccccc}
    \hline
    \textbf{Num. Images} & \textbf{32} & \textbf{64} & \textbf{128} &    \textbf{256} & \textbf{512} \\
    \hline
    Genie-M~\cite{Genie} & 16.17 & 33.13 & 42.29   & {48.37} &  51.50 \\
    MetaAug (Ours) & 23.79 & 37.71 & 44.25 & 49.07 & 52.32 \\
    \hline
  \end{tabular}
  \label{tab:less_data}
\end{table}
\section{Efficacy for various calibration data sizes} 
We validate the effectiveness of our proposed method using various calibration data sizes, from 32 to 512 images. Table~\ref{tab:less_data} shows that our method consistently outperforms Genie-M~\cite{Genie}, and the larger improvements are achieved with smaller calibration data sizes, e.g., the improvements are 7.62\% and 4.58\% with 32 and 64 calibration images, respectively. This demonstrates the effectiveness of our proposed method, especially in challenging conditions with limited data.

\section{More visualization as Fig. 1 in the main paper}
\begin{figure}[t]
        \centering
              \begin{subfigure}[b]{0.24\textwidth}
                 \centering
                 \includegraphics[width=\textwidth]{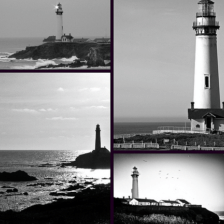}
             \end{subfigure}
             \hfill
            \begin{subfigure}[b]{0.24\textwidth}
                 \centering
                 \includegraphics[width=\textwidth]{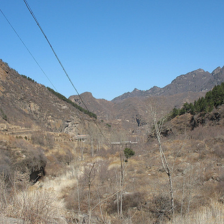}
           
             \end{subfigure}
            \hfill
            \begin{subfigure}[b]{0.24\textwidth}
                 \centering
                 \includegraphics[width=\textwidth]{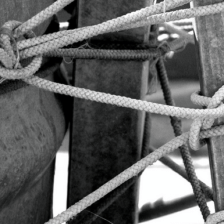}
          
             \end{subfigure}
             \hfill
             \begin{subfigure}[b]{0.24\textwidth}
                 \centering
                 \includegraphics[width=\textwidth]{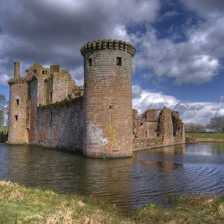 }
             \end{subfigure}
            \hfill

            \begin{subfigure}[b]{0.24\textwidth}
                \centering
                \includegraphics[width=\textwidth]{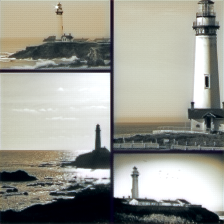}
            \end{subfigure}
            \hfill
           \begin{subfigure}[b]{0.24\textwidth}
                \centering
                \includegraphics[width=\textwidth]{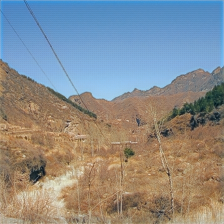}
          
            \end{subfigure}
           \hfill
           \begin{subfigure}[b]{0.24\textwidth}
                \centering
                \includegraphics[width=\textwidth]{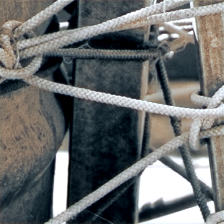}

             \end{subfigure}
             \hfill
            \begin{subfigure}[b]{0.24\textwidth}
                \centering
                \includegraphics[width=\textwidth]{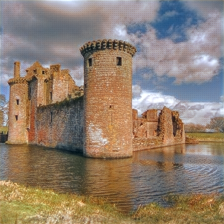}
            \end{subfigure}
           \hfill

           \begin{subfigure}[b]{0.24\textwidth}
            \centering
            \includegraphics[width=\textwidth]{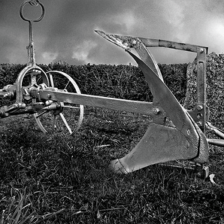}
        \end{subfigure}
        \hfill
       \begin{subfigure}[b]{0.24\textwidth}
            \centering
            \includegraphics[width=\textwidth]{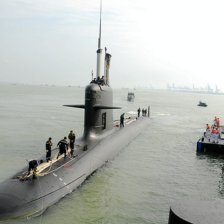}
      
        \end{subfigure}
       \hfill
       \begin{subfigure}[b]{0.24\textwidth}
            \centering
            \includegraphics[width=\textwidth]{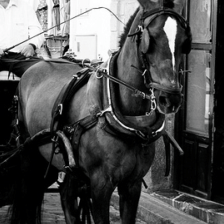}
     
        \end{subfigure}
        \hfill
        \begin{subfigure}[b]{0.24\textwidth}
            \centering
            \includegraphics[width=\textwidth]{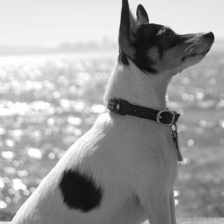 }
        \end{subfigure}
       \hfill

       \begin{subfigure}[b]{0.24\textwidth}
           \centering
           \includegraphics[width=\textwidth]{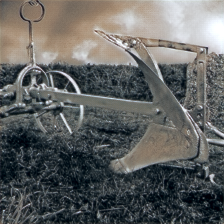}
       \end{subfigure}
       \hfill
      \begin{subfigure}[b]{0.24\textwidth}
           \centering
           \includegraphics[width=\textwidth]{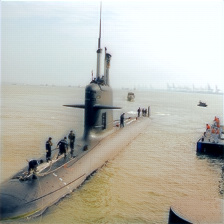}
     
       \end{subfigure}
      \hfill
      \begin{subfigure}[b]{0.24\textwidth}
           \centering
           \includegraphics[width=\textwidth]{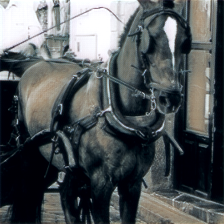}

        \end{subfigure}
        \hfill
       \begin{subfigure}[b]{0.24\textwidth}
           \centering
           \includegraphics[width=\textwidth]{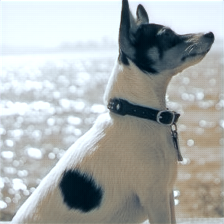}
       \end{subfigure}
      \hfill  
            \caption{Visualization of the original calibration images (the first and third rows) and the corresponding modified images (the second and fourth rows) produced by the transformation network.}  
            \label{fig:visualization}
        \end{figure}

\cref{fig:visualization} shows the visualization of the original images and the modified images using the proposed MetaAug. The results show that the modified images change the appearance of the original images while still preserving the semantic information of the original images.

\end{document}